\documentclass[11pt]{article}

\usepackage{amssymb,latexsym,amsmath,amsbsy}
\usepackage[dvips]{graphicx}
\usepackage{cite}
\usepackage{hyperref}
\newtheorem{theorem}{Theorem}

\newcommand{\BE}{\begin{equation}}
\newcommand{\EE}{\end{equation}}
\newcommand{\BQ}{\begin{equation} \begin{array}{c}}
\newcommand{\EQ}{\end{array}\end{equation}}
\newcommand{\BT}{\begin{theorem}}
\newcommand{\ET}{\end{theorem}}

\newcommand{\BL}{\begin{lstlisting}}
\newcommand{\EL}{\end{lstlisting}}

\newcommand{\GG}{\Gamma}
\newcommand{\PB}{\widehat{P}}
\newcommand{\YB}{\widehat{Y}}
\newcommand{\LAG}{\mathcal{L}}
\newcommand{\PP}{\mathcal{P}}

\newcommand{\DD}{\mathcal{D}}

\usepackage{listings}
\usepackage{color}

\begin{document}

\begin{center}
\noindent
{\Large \bf
Connections between physics, mathematics and deep learning.
}
\\

\vskip 10mm

{\bf Jean Thierry-Mieg}\\[2mm]
NCBI, National Library of Medicine, National Institutes of Health,\\
8600 Rockville Pike, Bethesda MD20894, USA.\\
 E-mail: mieg@ncbi.nlm.nih.gov \\[2mm]
\end{center}





\vskip 10mm
\begin{abstract}
Starting from the Fermat's principle of least action, which governs
classical and quantum mechanics and from the theory of
exterior differential forms, which governs the geometry of curved manifolds,
we show how to derive the equations governing neural networks
in an intrinsic, coordinate-invariant way, where
the loss function plays the role of the Hamiltonian.
To be covariant, these equations imply a layer 
metric which is instrumental in pretraining 
and explains the role of conjugation when using complex numbers.
The differential formalism clarifies
the relation of the gradient descent optimizer
with Aristotelian and Newtonian mechanics.
The Bayesian paradigm is then analyzed as a renormalizable
theory yielding a new derivation of the
Bayesian information criterion.
We hope that this formal presentation of the
differential geometry of neural networks will encourage some physicists 
to dive into deep learning, and reciprocally, that the specialists of
deep learning
will better appreciate the close interconnection of their subject
with the foundations of classical and quantum field theory.

\end{abstract}
\vskip 10mm

\section{Background}
\subsection{What is a neural-net good for}
The purpose of a neural-network, the logical architecture behind deep
learning \cite{[1]}, is to transform an input vector $X$ 
into a labeling vector $\YB$, for example,
in a supervised learning problem, the vector $X$ may represent
an image and the output $\YB$ a classification probability
like 'this image has a probability 92/100 of representing a cat'.
In more complex settings like reinforcement learning and adversarial
networks \cite{[2],[3]}, the $\YB$ may represent a choice between several actions.
But in all cases, during the training phase, one provides a set of $N$ training
vectors $\{X\}$, computes the corresponding set of output
vectors $\{\YB\}$ and compares $\{\YB\}$ to a set of truth vectors
$\{Y\}$ representing the desired outputs of the neural net. The comparison is performed
by selecting a loss function 
  
\BE
\LAG = \frac {1}{N} \; \sum_{X} \LAG(X). 
\EE
$\LAG$ plays the role of the Hamiltonian in classical mechanics and will be
 used to define the time-flow of the neural network during the training iterations.

\subsection{How is it designed}
A neural net consists of a large collection of very simple 
interconnected computing cells  described below, called artificial neurons. Each
neuron acts nearly trivially, but complex learning emerges from their combination.  
The actual design of a neural-net is specified by a set of choices, called the hyper-parameters
of the net. They include the number of layers of the net, 
the types and number of neurons in each layer, and their connectivity.
The hyper-parameters are selected by the user and are not automatically adjustable.
At present, the design of the network remains an art, but it has been observed that deep
nets, with many layers, learn better than shallow nets as each successive layer
spontaneously builds a higher-level representation of the data.
In 2017, many publications
described neural-net with over 100 layers.
Hence the re-branding around 2007 of the 'Artificial Neural Nets' paradigm of the 80's 
under the new name 'Deep learning'.

\subsection{How does it work}
The magic of the neural-net is that during the training phase, the neural-net
automatically modifies the response of its individual neurons
until the predicted outputs $\{\YB\}$ closely match the desired outputs $\{Y\}$. 
Here is how it works:

Each neuron performs an affine transformation $Y = WX + b$ on its input vector $X$,
followed by a nonlinear activation function $\Phi$ like a sigmoid, or a
rectified linear unit (ReLU), only allowing the propagation of positive
signals ($Y > 0$). The  $Z = \Phi(Y)$ are then used as the $X$ input
of the next layer. The activation functions play a  crucial role. Without them,
the network would be equivalent to a large matrix multiplication and could
only solve a linear classification problem. But the presence of several layers 
of non-linearities allows the neural-net to
learn to recognize very complex relationships in the input data, up to understanding
natural language, playing chess and go, or driving a car.
The basic method is amazingly simple.

At time zero, the $W$ and $b$ coefficients, collectively 
called the parameters of the neural-net, are initialized with small random numbers
in order to break any existing symmetry.
Then the time evolution of the parameters is driven by a simple
differential equation, called the steepest descent, which depends on
the choice of the loss function $\LAG$.

To construct this equation, 
the main idea is to realize that $\LAG$ is not supposed to constrain 
the $\{X\}$ vectors, which represent the external data, for
example texts or images, but should rather be regarded as a function of the parameters
$\{W\}$ of the neural-net, and that this dependency survives the 
averaging over the $X$
\BE
\LAG (W) = \frac {1}{N} \; \sum_{X} \LAG(X, W). 
\EE

We now remember that we want the $W$ to evolve in time
until the network is well-trained, so we postulate that the $W$ are unknown functions of time:
\BE 
W = W(t).
\EE
Hence the loss function itself becomes a function of time, hopefully converging
towards a global minimum:
\BE 
\LAG(t) = \LAG (W(t)).
\EE
Let us now compute the exterior differential of $\LAG$
\BE
d \LAG = \frac {\partial \LAG}{\partial t} \;dt = \sum_{W} \frac {\partial \LAG}{\partial W} \;\frac {\partial W}{\partial t} \;dt.
\EE

Starting from the differential of the loss function, it is then generally postulated \cite{[4]} that the 
time evolution of the $W$ parameters is governed by the differential equation
\BE
dW = - \frac {\partial \LAG}{\partial W} \;\eta \;dt,
\EE
where the parameter $\eta$ is called the learning rate.
The rationale for postulating this equation is explained in the next section
using the fundamental concepts of mathematics and physics.

\section{Neural nets from the point of view of differential geometry}
\subsection {Fermat's principle of least action}

The governing principle of a very large part of theoretical physics, including
general relativity, classical and quantum mechanics, and the standard model
of the fundamental interactions, is a suitable generalization of Fermat's principle
of least action \cite{[5]}. In its original form, in the seventeenth century, it simply stated
that a ray of light will follow the fastest path between two points, explaining refraction by
assuming that light travels slower in water than in the air, a true
statement which was verified experimentally only much later and was in plain
contradiction with the unfortunate hypothesis of Descartes that light would travel faster in water than in the air.
Around 1930, following Elie Cartan, Einstein and Hermann Weyl, 
it had become apparent that the best formalism to express
the least action principle is the formalism of exterior differential geometry,
whereby particles travel along straight lines, called geodesics, in a curved space
representing the presence of external forces, like electromagnetism or
gravity, and the eventual existence of constraints.
 
We would like to show that the training of a neural network
follows the same paradigm and can be expressed in the same formalism.
This is not really new or surprising, but this point of view is not 
emphasized in the recent book by
Goodfellow, Bengio and Courville \cite{[2]}, nor in the book of G\'eron \cite{[3]}, 
nor in the excellent lectures of Ng \cite{[6]}.

The neural-net steepest descent equation implements Fermat's principle of least action in the
following sense. The neural-net flows along the shortest path in parameters space leading 
to a given decrease of the loss function. However, to define a distance 
in $\{W\}$ space, we need a metric $g$. If we call $W^{[i]}$ an individual parameter, 
for example, a matrix coefficient associated to the $i^{th}$ layer, the neural-net equation reads:
\BE
dW^{[i]} = - g^{[ij]} \frac {\partial \LAG}{\partial W^{[j]}} \;\eta \;dt .
\EE

Notice the presence in this equation of upper and lower indices, respectively called
contravariant and covariant, which are needed each time one wishes to write
consistent equations in a system of coordinates which is not orthonormal, i.e. either
the axes are not orthogonal or the base vectors
have different lengths.
 
The layer metric $g$ is needed in two equations: it is needed to transform the
partial derivative $\partial_{[j]} \LAG$ with a lower (covariant) $[j]$ index into
a quantity with an upper (contravariant) $[i]$ index, so that it can be added to the
upper (contravariant) index differential $dW^{[i]}$. A metric is also needed 
to construct in  $\{W\}$ space the elementary square distance $ds^2$, 
familiar from general relativity \cite{[7]}: 
\BE
ds^2 = g_{[ij]} dW^{[i]} dW^{[j]},
\EE
where the lower (covariant) index metric $g_{[ij]}$ is defined as the inverse of the
upper (contravariant) index metric $g^{[ij]}$:
\BE
g^{[ij]} g_{[jk]} = \delta_{[k]}^{[i]}.
\EE
The parameters of each layer $[i]$ are naturally organized as a matrix $W^{[i]}$
feeding its output vector as the input of the following layer.
In this matrix notation, the $ds^2$ is expressed as the Frobenius norm:
\BE
ds^2 = g_{[ij]} \;\;Tr (dW^{[i]t} \;dW^{[j]}),
\EE
where the $t$ superscript denotes the transposed matrix.
By definition of the matrix product, this equation is just a covariant notation for 
the sum of the squares of all elements of the $dW$ matrix. weighted by the $g$ metric.
If we choose the same metric $g$ in equations (7) and (8),
the neural-net equation then implies that the length
of the path from an initial configuration $W_0$ to a final configuration $W_1$
computed as 
\BE
I = \int_{W_0}^{W_1} ds
\EE
is minimal relative to the distance from $W_0$ to any other (local) configuration $W$ 
with the same loss function as $W_1$, exactly as required by the principle of least action \cite{[8]}.
At each instant, the $W$ parameters flow normally to
the sheet of configurations with equal loss, where orthogonality is
defined relative to the metric $g$.

The existence of the layer metric $g$ is implied by the structure of the equations.
But from a pragmatic point of view, it plays a useful role. Its meaning
is that all cell layers do not have to be created equal. A classical method introduced
 by Hinton is called pretraining. One first trains a rather shallow neural-net on a large
 set of unlabelled examples, allowing the neural-net to recognize the main features of a 
new kind of data, and then one freezes the coefficients of these layers and trains 
additional layers which try to transform the output of the shallow network into the 
desired results using a possibly smaller set of labelled examples with known truth values.

 For example, suppose that the
first 6 layers of the network were pretrained and that 3 additional layers
need training. Using a diagonal metric, we would set
$g^{[ij]} = \delta^{ij}$ (the unit matrix), for $i,j > 6$
so that they would be trained normally. However,
we would set the upper index (contravariant) metric
$g^{[ij]}$ to zero for $i,j <= 6$, or equivalently
the lower index (covariant) metric $g_{[ii]}$ to $\infty$,
making the parameters of the low layers immutable.
Alternatively, we could set the low layer  $g_{[ij]}$ to
a high value, like $100$, allowing the pretrained part of the network to
adjust conservatively to the new condition at a very slow rate. We could also
decide that the parameters of layers with many cells are 
stiffer or softer than the parameters of layers with fewer cells.
Here we have treated the metric as layered, but if a layer contains several distinct
types of cells, it also makes sense to give a different stiffness  to each group.  Such
techniques are widely used when pretraining deep networks.

We see that writing the equations of the neural net in the classical
notations of the physicist forced us to introduce in (7) a metric which
was not apparent in (6) and to
anticipate the concept of variable stiffness of the successive layers of the neural-net.

An important observation is that thanks to the linearization procedure,
inherent to the differential formalism,
we never need to compute the inverse of a matrix. In the forward 
action we compose the successive actions of several layers.
In the pullback equation, which maps the differential $d\LAG$ of the loss function
back to the differential $dW$ of the parameters,
we only need the transpose of the Jacobians of the forward actions, not their inverse.
This is crucial, because a neural-net often involves very large matrices, and computing
the inverse of a very large matrix is at best very slow and very often 
numerically unstable.

\subsection {Understanding the metric when using complex numbers}

A way to illustrate the role of the $g$ metric is to analyze the 
situation when the $W$ coefficients are complex numbers. 
The square length of a complex number $z = x + iy$ is not given
by the square of $z$ but by the product of $z$ by its conjugate $\overline{z}$.
In other words, in $z,\; \overline{z}$ space, the metric is anti-diagonal
\BE
g_{z\;z} = g_{\overline{z}\;\overline{z}} = 0,\;\;\;\;\;\;\;\; g_{z\;\overline{z}} = g_{\overline{z}\;z} = 1/2\;. 
\EE
As a result, we find that the differential of $W$ is proportional to
the derivative of $\LAG$ with respect to $\overline{W}$ rather than with respect to $W$
because the $g$ metric in (7) will always couple a complex to its conjugate:
\BE
dW^{[i]} = - 2 \;g^{[ij]} \frac {\partial \LAG}{\partial \overline{W^{[j]}}} \;\eta \;dt .
\EE
The need to take the partial derivatives with respect to the complex conjugates of
the parameters would not be self-evident if we had not explicitly introduced the
$g$ metric. 

Complex neural networks are naturally important in domains where the input vectors $X$ are
best described by complex functions, as in sound recognition or imaging where the 
phase of the signal characterizes the direction of the source. But they are also 
promising  in other domains. The complex differentiable (holomorphic) functions are
much more constrained than real differentiable functions, and the space of
vectors of norm one ($z \overline{z} = 1$) is connected in the complex case,
the points of norm one $z = e^{i\phi}$ form a continuous circle, whereas the points
of norm one on the real line $x = \pm 1$ are disconnected.
 These two properties should facilitate the exploration
of the parameter landscape and although complex neural networks are not
yet natively supported in TensorFlow, we expect that they will be
widely used within a few years. See \cite{[9]} for a recent application of complex 
neural nets to the analysis of MRI medical pictures,
\cite{[10]} for an application to sound patterns, 
or \cite{[11]} for an introduction to the complex Cayley transform.

\subsection{Mechanical interpretation of the gradient descent optimizers}

The loss function $\LAG (W)$ can be interpreted as the potential energy 
of the system, usually denoted $V(x)$ in classical mechanics.
The negative of the gradient of $\LAG$ with respect to $W$ 
therefore represents the force $F$ causing the network to move across 
the parameter space $W$ with speed $v$. In these notations, the pullback equation
reads:
\BE
 v = \frac{\partial W}{\partial t} =  \eta F.
\EE
As in Aristotle's mechanics \cite{[12]}, this equation tells us that the speed $v$
of the mobile is proportional to the force. This equation is
physically correct only in a situation dominated by a huge friction, like
a horse pulling a plough. In those cases, the motion
is usually very slow. If we hope to accelerate the convergence of
the network, it seems reasonable to look for an equation
applicable to cases with less friction and faster
displacement and to postulate with Newton that
the acceleration $a$, rather than the speed $v$, is 
proportional to the force, according to the equation:
\BE
  m a = m \frac{\partial v}{\partial t} = F - \lambda v
\EE
describing the acceleration $a$, of a point of mass $m$, 
subject to a force $F$, with friction 
coefficient $\lambda$. The mechanical inertia
associated to the mass of the mobile stabilizes the
module of the speed and the orientation of the trajectory.
On a flat section of the landscape, where $F = 0$, the motion continues and 
the speed $v$ only decays exponentially as $e^{-\lambda t/m}$.
This method, introduced in \cite{[13]}
is called the gradient descent 'momentum' optimizer.
As hoped, the network converges faster and more often
than with the Aristotle equation. 

The current best methods, RMS-propagation \cite{[14]} then Adam \cite {[15]},
introduce a further
refinement. Close examination of the trajectories shows that the network is subject 
to a Brownian motion because each new set of training examples
introduces a modification of the loss function $\LAG(W)$ and
tends to drive the weight configuration in a different direction \cite{[16]}. However, only the average 
motion is desirable. A solution is to compute
\BE
\frac {\partial \overline{F}}{\partial t} = - \beta (\overline{F} - F) 
\EE
which defines $\overline{F}$ as the rolling average of $F$, with exponential time decay $\beta$,
and to postulate the descent equation:
\BE
  m a = m \frac{\partial v}{\partial t} = \gamma \overline{F} - \lambda v
\EE
where the variable coefficient $\gamma$ dampens the effect of the
components of $\overline{F}$ in the directions in which $F$ fluctuates,
as measured by maintaining the rolling exponential time average 
of $F_w^2$ in each $w$ direction. 
These methods strongly accelerate the convergence towards a 
good local minimum of $\LAG$, although it is sometimes reported that
the network is over adapted to the examples and does not
generalize so well to new test examples \cite{[17]}.

\subsection {On the paucity of local minima in high dimension}

A network can only be trained well if the gradient descent
paradigm can discover configurations
with a very low loss function, such that each
training example ${X}$ is mapped very close to its
known target value $Y$. Furthermore, one hopes that
such a good mapping will generalize well to
new test examples not seen during the training. Therefore, a very interesting question 
is to evaluate the risk of 
being trapped in a false minimum. 

Drawing from our life-long 3-dimensional experience, we expect local minima
to be very frequent: in a mountain landscape, there are
many lakes and on a rainy day huge numbers of little puddles of water are forming.
However, neural networks often have millions of
$W$ parameters, and, in high dimension, local minima become extremely rare
relative to saddle points \cite{[1]}. In a space of dimension $D+1$
a horizontal plane tangent to an equipotential $\LAG$ surface
is defined by $D$ linear equations, indicating that each partial derivative
relative to a different direction vanishes. In each of these directions,
the second derivative may point up or down, yielding $2^D$
configurations, but only one of them, when all second derivatives point upwards,
corresponds to a local minimum. All other configurations
characterize saddle points where some escape routes remain open.
The true local minima are therefore exponentially rare, with probability
$2^{-D}$, relative to the saddle points, and this helps to understand why 
neural nets are not constantly trapped in false minima.
Some authors even try
to show that, in concrete situations, the different
minima discovered in the network are most often
connected by a quasi-horizontal path \cite{[18]}. These qualitative
observations may help understand the otherwise amazing success
of gradient descent equation to find deep minima
in these extremely complex manifolds. It would be interesting
to know in which sense the conjecture that there
would exist a single connected globally minimal region
could be validated. 

\subsection {Finite learning steps}

On a computer, we can only deal with a finite number of steps of calculation,
so we must replace the infinitesimal differential equation (6)
by the approximate finite difference equation
\BE
\delta W = - \frac {\partial \LAG}{\partial W} \;\eta \;\delta t,
\EE
where $\eta$ is the learning rate and $\eta\delta t$ now represents a small 
but finite quantity called the learning step.
If the step is too small, one
needs too many iterations, if too big, the linearization approximation
may be broken since some terms of order $(\eta \delta t)^2$ may become
as large or larger than some terms linear in  $\eta \delta t$. These 
non-linearities interfere with the logic of the calculation which may 
become unstable and miss the true minimum. Of course, following
the classical Runge-Kutta methods dating back to 1900, it is recommended to adapt the step
to the steepness of the differential equation and go fast in shallow
regions and slow over cliffs. However, it
must be understood that the main cause of the problem
is not the excessive step $\delta W$ in one of the $D$ directions,
where $D$ is the number of parameters, but the
possible interferences between the $D^2/2$ pairs
of variables, the $D^3/6$ triplets, and so on,
interferences which do not exist in the
truly infinitesimal $dW$ formalism. 
The problem is well illustrated by the model of a car driving
on a multi-lanes freeway. Using differential equations, the car may
continuously adapt its direction and follow its own lane, 
but if it moves by quantum jumps, it may well in a
bend change lane and end up on an exit ramp, away from
its final destination.

The finite learning steps  have however two advantages.
First, they allow the introduction of activation functions,
like the ReLU diode, presenting (non-differentiable) angles because the
difference equation (18) remains well defined.
At the same time, they allow the network to traverse
the thin ridges  and to jump over the narrow ditches
which may be present in the parameter landscape \cite{[16]}.
This type of evolution is analogous to the tunnel effect
which allows electrons to traverse transistors
and prompts us to sketch the quantum mechanical
aspects of the theory of neural-networks.

\section {Renormalization theory and Bayesian Statistics}

Up to here, we have shown how neural nets are governed by the principles
of classical mechanics. In this chapter, following a suggestion
of the referee, we draw the correspondence between Bayesian 
statistics and modern quantum field theory and show how the
renormalization procedure helps answering a practical
question: how many parameters can be trained given the number of
training examples.

\subsection {The cross-entropy loss function}

To make the connection with thermodynamics and quantum mechanics, we must first
revisit the definition of the loss function $\LAG(W)$.
Given numerical outputs $\{\YB\}$, the simplest
way to compare them to the desired results $\{Y\}$ is to
choose as loss function the Euclidean distance
\BE
  \LAG = \frac {1}{2} \;\sum\; (Y - \YB)^2
\EE
where the sum extends over all the training examples.
The gradient of $\LAG$ is proportional to
the difference $(Y - \YB)$
\BE
  d\LAG = \sum_a\;(Y - \YB)_a\; \frac {\partial \YB^a}{\partial W^i}\;dW^i
\EE
and the gradient descent equation (7) is simple.

When the desired outputs are
qualitative, as in a classification problem,
a more complex loss function, called
the cross-entropy, is used. To understand its definition,
assume that the ${W}$ parameters are
known and compute the probability of correctly assigning each
${X}$ example to its correct class, i.e. the probability of a 
perfect classification
\BE
\PP = \prod_a (\PB^a)^{n_a},
\EE
where $\PB^a$ is the probability of correctly assigning an example belonging to class $a$
and $n_a$ is the true number of training examples belonging to class $a$.
Notice that there is a single exact configuration, so there is no need for
a combinatorial factor. The log of $\PP$ becomes a sum over all classes
\BE
  log (\PP) = \sum_a n_a\;log(\PB^a).
\EE
 In the limit where $n_a$ is a large number, $n_a$ converges
to $N P_a$, where $N$ denotes the total number of training samples
and $P_a$ denotes the true probability of class $a$ in the training set.
The log of the probability therefore converges towards
\BE
  lim_{N \rightarrow \infty} \;log(\PP) = N \;\sum_a P_a log (\PB^a).
\EE
Since all the probabilities are smaller than 1, the logs are negative,
and it is more natural to insert a minus sign and define the quantity
\BE
\LAG_0 = - \;\;\sum_a\;P_a\;\;log(\PB^a).
\EE
$\LAG_0$ is called the cross-entropy, it measures the distance between the desired probability distribution
$P$ and the predicted distribution $\PB$ and is closely related to the Shannon entropy $- \sum\;P\;log(P)$.
To construct a probability distribution from the output vector $\YB^a$ of the
neural network, one postulates a Boltzmann like distribution, called the soft-max:
\BE
\PB^a = \frac {e^{\YB^a}}{\sum_b {e^{\YB^b}}}\;\;\;\;\Rightarrow\;\;\;\sum_a {\PB^a} = 1
\EE
where the $e^{\YB}$ play the role of the familiar  energy/temperature ratios  $e^{-E/kT}$. As usual,
the zero-energy level is arbitrary: the probabilities $\PB^a$ are not modified if all the $\YB$ are shifted by 
the same constant.

Despite the apparent complexity of these definitions, this choice is magic.
As can be verified by a direct calculation, the gradient equation
\BE
 d\LAG = \sum_a\;(P - \PB)_a \; \frac {\partial \YB^a}{\partial W^i}\;dW^i
\EE
is nearly identical to the Euclidean gradient equation (20), with the simple
replacement of the difference $Y - \YB$ by $P - \PB$. This
extremely beautiful equation is one of the jewels
of the back-propagation algorithm.

\subsection {The Bayesian integral}

Using classical statistics, we have computed (21) the probability of a perfect
classification, given a model specified by a set of parameters $\{W\}$.
However, since the network is only trained on a finite number of examples $N$,
the parameters of the optimal model $\{W_0\}$ cannot be known exactly,
and the crucial question is to estimate if the network will generalize well
in future tests or if it is over-fitting the training set.
As a proxy, we propose to estimate the probability of
a perfect classification if we consider a family 
of models approximating the unknown optimum.
Using the Bayes formula, this quantity can be expressed
as the product of $\PP$, which represent the conditioanl probability of
a the perfect classification given a choice of the $\{W\}$, by the prior
probability $P(W)$ of the $\{W\}$ configuration, summed over all $\{W\}$
configurations. This sum is expressed by the integral
\BE
\GG_N = \int dW \; P(W)\; \PP(W) = \int dW\;  P(W)\; e^ {log(\PP(W))} = \int dW\;  P(W)\; e^ {- N \LAG_0(W)}
\EE
where $\LAG_0$ is the cross-entropy loss function (24).
In the absence of any prior knowledge, the sum over $W$ is unbounded
and the $\GG$ integral diverges.

\subsection {Regularization}
Facing a divergent integral is familiar in quantum field theory. 
The canonical way to work around this difficulty 
may seem artificial and counter-intuitive,
but it is validated by innumerable accurate experimental predictions
in statistical and particle physics. 
It involves two steps. In the first step, called regularization,
a regularizer $\lambda$ is introduced
such that all integrals converge when $\lambda$ is finite.
In the second step, called renormalization, 
one tries to construct quantities 
which converge towards a finite value
when the regularizer goes to infinity.
Only these finite limits are called the observables of the theory.

In the present situation, to insure the convergence of the
intermediate calculations, we limit the range of variation of the parameters
by supposing that the prior probability $P(W)$ can be represented by
a Gaussian distribution with large variance $\lambda^2$ and k-dimensional volume 1:

\BE
\GG_0 = \int dW P(W) \;=\; \int dW (\frac {1}{\sqrt{2 \pi \lambda^2}})^{k} exp (\sum_{i=1}^k\; - \frac{W_i^2}{2 \lambda^2}) \;\;= \;\;1,
\EE
where $k$ denotes the number of parameters, i.e. the dimension of the $W$ space. The Gaussian factor
$-(W_i)^2/2 \lambda^2$ can be written in a covariant way as $- g_{ij}\;W^iW^j/2$ where the metric $g_{ij}$ is 
$1/\lambda^2$ times the $k$-dimensional identity matrix $\delta_{ij}$. Its determinant $g$ is equal 
to the product $\lambda^{-2k}$, therefore we can rewrite $\GG_0$
in a covariant way as
\BE
\GG_0 = \int dW \;\sqrt{g}\;\; (2 \pi)^{-k/2} exp (- g_{ij} W^iW^j/2).
\EE
We recognize $ d W \;\sqrt{g}$ as the covariant Riemannian volume element  \cite{[5],[7],[8]}.

Let us now define the regularized loss function
\BE
\LAG_1 = \LAG_0 + \sum_i \frac{W_i^2}{2 N \lambda^2},
\EE
where $\LAG_0$ is the cross-entropy defined in (24) and $N$ is the number of examples. 
$\LAG_1$ tends to $\LAG_0$ when $N\lambda^2$ tends to infinity.
Substituting (29,30) in (27), we obtain
\BE
\GG_N = \int dW \; \;\sqrt{g}\;\; (2 \pi)^{-k/2} \; e^{-N\;\LAG_1(W)},
\EE
which is similar to the thermodynamics partition functions
\BE
 \int \DD\phi\;  P(\phi)\; e^ {-(1/kT)  \int dx H(\phi(x))}\;,
\EE
where $H$ is the Hamiltonian of the systems.
We learn in this way that $1/N$ plays the role of the absolute temperature $kT$ 
and is the natural parameter to use in a perturbation expansion.

It should be noticed that in neural network applications \cite {[2],[3],[6]}, a corrective factor $W^2/2 \alpha^2$ 
is often added to the loss function to limit the range of the $W$ parameters during learning. We learned
in (30) that the coefficient $\alpha^2$ should scale like $N \lambda^2$ and we realized that to be rigorous
we need a compensating term $(2\pi \alpha^2/N)^{-k/2}$ in the measure (28).

\subsection {The large number hypothesis and the regularized ground state}

Let us now make the bold supposition that $\LAG_1(W)$ has a single global minimum at position $W_1$.
Call $w = W - W_1$ the displacement away from this extremum and develop $\LAG_1$ in a Taylor series
to second order in $w$. The terms linear in $w$ vanish, since we are at an extremum and we can write
\BE
\begin{array}{ccccccc}
  \LAG_1(W)  = \LAG_1(W_1) + \frac {1}{2} h_{ij} w^i w^j,
\end{array}
\EE
where $h_{ij}$ denotes the matrix of the second partial derivative 
\BE
  h_{ij} = \frac {\partial^2  \LAG_0}{\partial w^i\;\partial w^j} +  \frac {\delta_{ij}}{N \lambda^2}\,.
\EE
In geometry, $h_{ij}$ is called the Hessian and characterizes the curvature radii
of the k-dimensional ellipsoid best contacting the surface $\LAG_1$ in the vicinity of $W_1$.
In statistics, the Hessian of $\LAG_0$ is called the Fisher information
matrix for a single data point and the second order Taylor expansion of $\LAG_1$ 
can be seen as an application of the law of large numbers because the variance of the
Gaussian distribution $exp(-N\;\LAG_1)$ (31) is of order $1/N$ which becomes very narrow when $N$ is large.
Using (29) to evaluate the integral of the Gaussian $exp (- N\;h_{ij} w^iw^j/2)$, we obtain
\BE
\int dw P(W_1 + w) e^{-N \LAG_1(W_1+w)} = N^{k/2} \frac {\sqrt{g}}{\sqrt{h}} e^{-N \LAG_1(W_1)} = e^{-N \LAG_2(W_1)}
\EE
where
\BE
\LAG_2(W_1) = \LAG_1(W_1) + \frac {k}{2}\;\frac{log(N)}{N} + \frac {1}{N} log (\frac{\sqrt{h}}{\sqrt{g}})
\EE
where $g$ is the metric of the vacuum (29), $h$ is the determinant of the regularized Hessian (34), 
$W_1$ is the position of the minimum of $\LAG_1$  and the factor $(2 \pi)^{k/2}$ present in (29) has canceled out.
The dependency in $N$ is not fully explicit. First, the position $W_1$ of the minimum of $N \LAG_1$
is shifted relative to the position $W_0$ of the minimum of $N \LAG_0$ by the presence of the regularizing
term $W^2/2\lambda^2$ (30), contributing a correction of order $1/N \lambda^2$. Furthermore, the Hessian $h_{ij}$
differs  from the Fisher matrix. Assuming an orthonormal frame, each eigenvalue
(34) is shifted from $1/\sigma^2$ to $1/\sigma^2 + 1/N\lambda^2$, where $1/\sigma^2$ is one of the
eigenvalues of the Fisher matrix,
contributing a second correction of order $1/N \lambda^2$. In a general frame, the correction
is more complex, but still of order $1/N \lambda^2$ so the sum of the 2 corrections 
can be written as $\LAG_2(W_1) = \LAG_2(W_0) + c/N \lambda^2$ and  vanishes when $N$ or $\lambda^2$ go to infinity.
Notice however that $\LAG_2(W_1)$ (36) does not correspond to an observable: it diverges when $\lambda$ goes to infinity since $\sqrt{g}$ goes to zero and therefore $log(\sqrt{g})$ diverges. 

\subsection {Renormalization}

To extract an observable from the regularized minimum $\LAG_2(W_0)$, the term $log(\sqrt{g})$
must disappear. This happens if we compute the difference between two models
\BE
\Delta \LAG_2 =  \Delta \LAG_0 + \frac {\Delta(W_0^2)}{2N\lambda^2} + \frac {\Delta k}{2}\,\frac {log(N)}{N}
+ \frac{1}{N} \;\Delta log(\sqrt{h}) + \frac {\Delta c}{N \lambda^2}.
\EE
This quantity is renormalizable. When $\lambda$ goes to infinity, the regularizing term $\Delta\;W_0^2/2N\lambda^2$ and the correction factor $\Delta c/N\lambda^2$ both tend to zero even if $N$ is kept finite and we obtain
the finite observable
\BE
   \Delta \LAG_2 = \Delta \LAG_0 \;+\; \frac {\Delta k}{2}\,\frac {log(N)}{N} + \frac{1}{N} \; log(\frac{\sqrt{h_1}}{\sqrt{h_2}}).
\EE
The advantage of the covariant $\sqrt{h}$ notation is that it is valid in any system of coordinates and we do not have
to assume that $h_1$ and $h_2$, the Hessian or Fisher Information matrices of the two families of models, can be diagonalized at
the same time.

The quantity $(2N\;\Delta \LAG_0 \;+\; \Delta k\;log(N))$ is called the Bayesian Information 
Criterion (BIC) \cite{[19]}. It quantifies a natural idea:
when the number N of training examples is large, we can train a large network
with a large number of parameters $k$, but if $N$ is limited
we cannot. The network would over-fit the training set 
and not generalize properly to the test set.
The BIC factor indicates that the sweet number of parameters scales like $2N/log(N)$.

When $N$ is large, the third term in (38) is smaller. It is of order $1/N$ and proportional to the log of
the ratio of the volume elements of the two families of models.
This tells us that for a fixed value of the BIC, a family of models with a 
fancier dependency on the choice of the parameters, yielding a larger volume element, should
be penalized relative to a simpler model. This concept is well presented in \cite{[20]}.

In practical terms, when designing a neural network, we have found an evaluation of the number $N$ of examples 
needed to train $k$ parameters and we have
shown that the $L_2$ regularizer should scale like $1/N$.

It is also essential to understand that only the difference $\Delta \LAG_2$ (38) 
between two families of model is well defined. In a way, the renormalization effect $\Delta k \;log(N)/2N$ 
is analogous to the  Casimir effect, predicted in 1948 \cite{[21]} and experimentally verified in 1997 \cite{[22]}. 
Two neighboring conducting plates attract each other, even in the absence of electric charges,
because the pressure from the electromagnetic fluctuations of the
vacuum existing outside the capacitor are not fully compensated by the fluctuations existing between the 
plates, since there is not enough room to allow long wave-length vacuum fluctuations in this narrow space. 
The effect is small but can be measured \cite{[22]}. In the same way,
when we subtract the two regularized Bayesian integrals (37), the correction term comes from the fact that
a model with $k$ parameters cancels the long wavelength fluctuations of theses parameters, which are now
squeezed in a Gaussian of width $\sigma/\sqrt{N}$, whereas the remaining parameters fluctuate up to
the long wavelength $\lambda$. We cannot count the total number of virtual parameters, but we can
accurately estimate the influence of the removal of $\Delta k$ large fluctuations in the larger model. The metric term
corresponds to the calculation of the Casimir effect when the parallel planar conductors are replaced by a
capacitor with a more complex shape.

It would also be the proper formalism to consider gauge invariance with respect
to groups of transformations like translations and rotations of training images.
The equivalent of the Yang-Mills differential forms would be introduced in
the loss function $\LAG$ and since they are 1-forms, their pullback would
naturally trickle down into the descent equation.

\subsection {Looping the loop: the co-gradient analytic descent equation}

Training by gradient descent and Bayesian inference are usually considered
as distinct. However, the metric in parameter space, equations (7) and (34),
provides an operational unification. We illustrate this in the simple
case of a quadratic potential where the unified formalism provides a 
straightforward construction of the optimal single-step co-gradient descent.

The Bayesian formalism shows that, to second order, the natural metric of the $W$ space is the Hessian
of the loss function (34). If we
re-inject this choice in the covariant definition of the gradient descent (7), we obtain
the so called co-gradient descent. Assume that we are already in the vicinity
of the absolute minimum $W_0$, and that the loss function is truly quadratic in the $w = W - W_0$
\BE
\LAG = \frac{1}{2} a_{ij} w^i w^j.
\EE
The Hessian $h_{ij}$, i.e. the second derivative of $\LAG$,  is equal to the matrix $a_{ij}$, and 
since $h_{ij}$ is our choice for the metric $g_{ij}$ we have
\BE
\frac {\partial \LAG}{\partial w^i} = a_{ij} w^j = h_{ij} w^j = g_{ij} w^j.
\EE
The upper index metric $g^{ij}$ is the inverse lower index metric $g_{ij}$ (9), hence (7) simplifies to
\BE
    dw^k = - \eta \; dt\; g^{ki}\;\frac {\partial \LAG}{\partial w^i} = - \eta \; dt\; g^{ki}\;g_{ij}\; w^j = - (\eta\;dt) w^k.
\EE
Using a finite learning step $(\eta\;\delta t) = 1$ (18), $w$ jumps immediately to the true minimum
\BE
    w^k + \delta w^k = 0.
\EE
By reasoning on the differential geometric structure of the problem and on Bayesian probabilities
we have recovered the exact single step solution of this simple quadratic problem. 

In practice there are two limitations to this analytic method. The most obvious is that we cannot easily obtain
an accurate numerical estimation of the Hessian $h_{ij}$, and even if we could, computing its
inverse $h^{ij}$ would be numerically unstable when the number of parameters, i.e. the dimension
of the matrix $h$, is large. A second limitation is that the single step convergence does not depend on the 
sign of the second derivatives, more precisely on the sign of the eigenvalues of $h$. 
$\delta w^k$ can be oriented uphill as well as downhill towards a local extremum. The system would
therefore be trapped by the (very numerous) saddle-points. 
Thus, despite its beauty, the co-gradient method cannot be applied as is to deep learning.
Nevertheless, many gradient descent methods attempt to evaluate the Hessian in the direction of 
the propagation, i.e. the second difference
$w(t) - 2 w(t-1) + w(t-2)$, in order to adjust dynamically the learning rate $\eta$.

\section{Conclusion}
The purpose of this note was to clarify the training paradigm of
a neural net using the standard concepts and notations of 
differential geometry and classical mechanics,
a point of view not emphasized in the recent books
of Goodfellow, Bengio and Courville \cite{[2]}, or G\'eron \cite{[3]},
 or the lectures of Ng \cite{[6]}.
We have shown that the neural net steepest descent equation 
implements Fermat's principle of least action using the 
cotangent pullback of the differential of the loss function.
Since, as the name implies,
the cotangent pullback of a differential form uses the functions
describing each layer in reverse order, the back-propagation paradigm
of the neural-net is easily understood. We have also shown that to be covariant, 
the equations automatically imply a layer metric which is instrumental in
the pretraining of neural nets and
opens the possibility of working with all kinds of numbers.
In particular, if we use complex numbers, the metric
introduces an otherwise mysterious complex conjugation 
in the back-propagation equation. 
The mechanical interpretation of the loss function as the
potential energy of the network in parameter space helps
to understand why the 'momentum' method describes
a Newtonian system with less friction than the simple gradient descent
equation, and clarifies the Brownian motion aspects of the current best
optimizer, Adam. 

We also pointed out that the linearization procedure, 
implicit in any differential variation, 
avoids the calculation of inverse matrices,
greatly facilitating the implementation of the neural network  algorithms,
but that the finite steps $\delta t$ used on the
computer will break the linearization logic when $\delta t$ is too large
because some quadratic terms proportional to $\delta t^2$ may become larger
than some terms linear in $\delta t$. Finally, we recalled the beautiful
interplay between the Boltzmann distribution $exp(-E/kT)$ and the
choice of the cross-entropy loss function $-P \; log(\PB)$ leading to a gradient
directly proportional to $\PB - P$. 

We then showed that the Bayesian evaluation of the accuracy
of a neural network is given by an integral similar to
the partition function of thermodynamics. 
This Bayesian integral diverges, but it can be regularized
and the relative accuracy of two designs is renormalizable,
linking the number $N$ of training examples to the number $k$ of 
adjustable parameters of the network $k \sim 2N/log(N)$. It also follows
that the most natural choice for the metric
which appears in the gradient descent equation is
the Hessian of the cross-entropy loss function, called in statistics
the Fisher Information matrix, and that using this metric,
we recover the optimal co-gradient descent formalism.
Our introduction of the metric in paramater space (7) has provided 
a unification of training by gradient descent and Bayesian inference
which are usually considered as distinct problems.

We hope that this formal presentation of the
differential geometry of the neural networks will help some physicists 
to dive into deep learning, and reciprocally, that the specialists of
deep learning with a background in biology or computer science
will better appreciate the close interconnection of their subject
with the very rich literature on classical and quantum field theory,
in the hope that some of the latter techniques are still awaiting to
be transposed into Deep Learning.


\section*{Acknowledgments}

We would like to thank the referee who prompted the analysis of the
Bayesian formalism, Danielle Thierry-Mieg and Dominik Miketa
for critical suggestions, Mehmet Kayaalp and John Spouge for insightful 
discussions, and David Landsman for actively promoting research on neural networks.
This research was supported by the Intramural Research Program of the National Library of Medicine, National Institute of Health.


\section *{Appendix: Two simple analytic applications}

When we look at the computer programs used to train the neural networks, it may
seem that they work because they use successive discrete training
calculations. We show here, that the neural-net approach to equilibrium
follows normal differential equations which, in the simplest cases, can be integrated 
analytically using the usual rules of calculus. 

The first example is often used in neural-net as a
regulator. It corresponds to a mass term in classical mechanics.
To ensure the existence of a single global minimum, the loss function should be chosen to 
be convex and bounded from below. The simplest such function is the parabola,
\BE
\LAG (W) = \frac{1}{2} W^2
\EE
We have
\BE
d \LAG = \frac {\partial \LAG}{\partial W} \;dW = W \;dW,
\EE
hence the neural-net differential equation can be integrated analytically 
\BE
\begin{array}{ccccccc}
dW = - W \;\eta dt,
\\
\frac {dW}{W} = d (Log(W)) = - \eta dt, 
\\
W(t) = W_0 e^{- \eta t} ,
\\
\LAG(t) = \LAG_0 e^{- 2 \eta t}.
\end{array} 
\EE
The parameter of the net moves continuously down the parabola, and the
loss function decreases to zero in an exponential way.

The next simplest case is the quartic equation,
which illustrates the fact that a softer loss function slows down
the approach to equilibrium. Let us have:
\BE
\begin{array}{ccccccc}
\LAG(t) = \frac {\alpha}{8} W^4(t),
\\
d \LAG = \frac {\alpha}{2} W^3(t) dt.
\end{array} 
\EE
The neural-net differential equation becomes
\BE
\begin{array}{ccccccc}
dW =  - \frac {\alpha}{2} W^3 \;\eta \;dt
\\
- 2 \frac {dW}{W^3} = d \frac{1}{W^2} = \alpha \eta dt
\\
\frac {1}{W^2} - \frac {1} {W_0^2} = \alpha \eta t
\\
\\
W(t) = \frac {W_0} {\sqrt{1 + \alpha \eta W_0^2t}}
\\
\LAG(t) = \frac {\LAG_0} {(1 + \alpha \eta W_0^2 t)^2}
\end{array} 
\EE
where $W_0$ is the arbitrary initial value of the parameter. 
As expected the approach to equilibrium in $t^{-2}$ is slower than in the previous case,
which behaved as $e^{-t}$.

\end {document}